\documentclass[runningheads]{llncs}

\usepackage{graphicx}
\usepackage{adjustbox}
\usepackage{multirow}
\usepackage{array}
\usepackage{wrapfig}

\usepackage[T1,T2A]{fontenc}
\usepackage[utf8]{inputenc}
\usepackage[russian,english]{babel}
\usepackage{siunitx}
\newcolumntype{C}[1]{>{\centering}p{#1}}
\newcolumntype{R}[1]{>{\raggedleft}p{#1}}
\newcolumntype{L}[1]{>{\raggedright}p{#1}}

\begin{document}

\title{Dataset for Automatic Summarization of Russian News}

\author{Ilya Gusev\orcidID{0000-0002-8930-729X}}

\institute{Moscow Institute of Physics and Technology, Moscow, Russia\\
\email{ilya.gusev@phystech.edu}\\}

\maketitle

\begin{abstract}
Automatic text summarization has been studied in a variety of domains and languages. However, this does not hold for the Russian language. To overcome this issue, we present Gazeta, the first dataset for summarization of Russian news. We describe the properties of this dataset and benchmark several extractive and abstractive models. We demonstrate that the dataset is a valid task for methods of text summarization for Russian. Additionally, we prove the pretrained mBART model to be useful for Russian text summarization.

\end{abstract}

\section{Introduction}
Text summarization is the task of creating a shorter version of a document that captures essential information. Methods of automatic text summarization can be extractive or abstractive.

Extractive methods copy chunks of original documents to form a summary. In this case, the task usually reduces to tagging words or sentences. The resulting summary will be grammatically coherent, especially in the case of sentence copying. However, this is not enough for high-quality summarization as a good summary should paraphrase and generalize an original text.

Recent advances in the field are usually utilizing abstractive models to get better summaries. These models can generate new words that do not exist in original texts. It allows them to compress text in a better way via sentence fusion and paraphrasing.

Before the dominance of sequence-to-sequence models~\cite{sutskever_seq2seq}, the most common approach was extractive.

The approach's design allows us to use classic machine learning methods~\cite{wong_extractive_classic_ml}, various neural network architectures such as RNNs~\cite{lstm,summarunner} or Transformers~\cite{transformers}, and pretrained models such as BERT~\cite{bert,presumm}. The approach can still be useful on some datasets, but modern abstractive methods outperform extractive ones on CNN/DailyMail dataset since Pointer-Generators~\cite{pg}. Various pretraining tasks such as MLM (masked language model) and NSP (next sentence prediction) used in BERT~\cite{bert} or denoising autoencoding used in BART~\cite{bart} allow models to incorporate rich language knowledge to understand original documents and generate grammatically correct and reasonable summaries.

In recent years, many novel text summarization datasets have been revealed. XSum~\cite{xsum} focuses on very abstractive summaries; Newsroom~\cite{newsroom} has more than a million pairs; Multi-News~\cite{multinews} reintroduces multi-document summarization. However, datasets for any language other than English are still scarce. For Russian, there are only headline generation datasets such as RIA corpus~\cite{ria}. The main aim of this paper is to fix this situation by presenting a Russian summarization dataset and evaluating some of the existing methods on it.

Moreover, we adapted the mBART~\cite{mbart} model initially used for machine translation to the summarization task. The BART~\cite{bart} model was successfully used for text summarization on English datasets, so it is natural for mBART to handle the same task for all trained languages.

We believe that text summarization is a vital task for many news agencies and news aggregators. It is hard for humans to compose a good summary, so automation in this area will be useful for news editors and readers. Furthermore, text summarization is one of the benchmarks for general natural language understanding models.

Our contributions are as follows: we introduce the first Russian summarization dataset in the news domain\footnote{https://github.com/IlyaGusev/gazeta}. We benchmark extractive and abstractive methods on this dataset to inspire further work in the area. Finally, we adopt the mBART model to summarize Russian texts, and it achieves the best results of all benchmarked models\footnote{https://github.com/IlyaGusev/summarus}.

\section{Data}
\subsection{Source}
There are several requirements for a data source. First, we wanted news summaries as most of the datasets in English are in this domain. Second, these summaries should be human-generated. Third, no legal issues should exist with data and its publishing. The last requirement was hard to fulfill as many news agencies have explicit restrictions for publishing their data and tend not to reply to any letters.

Gazeta.ru was one of the agencies with explicit permission on their website to use their data for non-commercial purposes. Moreover, they have summaries for many of their articles.

There are also requirements for content of summaries. We do not want summaries to be fully extractive, as it would be a much easier task, and consequently, it would not be a good benchmark for abstractive models.

We collected texts, dates, URLs, titles, and summaries of all articles from the website's foundation to March 2020. We parsed summaries as the content of a ``meta'' tag with ``description'' property. A small percentage of all articles had a summary.

\subsection{Cleaning}
After the scraping, we did cleaning. We removed summaries with more than 85 words and less than 15 words, texts with more than 1500 words, pairs with less than 30\% unigram intersection, and more than 92\% unigram intersection. The examples outside these borders contained either fully extractive summaries or not summaries at all. Moreover, we removed all data earlier than the 1st of June 2010 because the meta tag texts were not news summaries. The complete code of a cleaning phase is available online with a raw version of the dataset.

\subsection{Statistics}
The resulting dataset consists of 63435 text-summary pairs. To form training, validation, and test datasets, these pairs were sorted by time. We define the first 52400 pairs as the training dataset, the proceeding 5265 pairs as the validation dataset, and the remaining 5770 pairs as the test dataset. It is still essential to randomly shuffle the training dataset before training any models to reduce time bias even more.

Statistics of the dataset can be seen in Table~\ref{tab1}. Summaries of the training part of the dataset are shorter on average than summaries of validation and test parts. We also provide statistics on lemmatized texts and summaries. We compute normal forms of words using the pymorphy2~\cite{pymorphy2}\footnote{https://github.com/kmike/pymorphy2} package. Numbers in the ``Common UL'' row show size of an intersection between lemmas' vocabularies of texts and summaries. These numbers are almost similar to numbers in the ``Unique lemmas'' row of summaries' columns. It means that almost all lemmas of the summaries are presented in original texts.

\begin{table}[htbp]
\caption{Dataset statistics after lowercasing}\label{tab1}
\footnotesize
\centering
\begin{tabular}{|c|S[table-format=7.1]|S[table-format=7.1]|S[table-format=7.1]|S[table-format=6.1]|S[table-format=7.1]|S[table-format=6.1]|}\hline
& \multicolumn{2}{c|}{Train} & \multicolumn{2}{c|}{Validation} & \multicolumn{2}{c|}{Test} \\\hline
& {Text} & {Summary} & {Text} & {Summary} & {Text} & {Summary} \\\hline
Dates & \multicolumn{2}{c|}{01.06.10 - 31.05.19} & \multicolumn{2}{c|}{01.06.19 - 30.09.19} & \multicolumn{2}{c|}{01.10.19 - 23.03.20} \\\hline
Pairs & \multicolumn{2}{S[table-format=7.1]|}{52400} 
&\multicolumn{2}{S[table-format=5.1]|}{5265} & \multicolumn{2}{S[table-format=5.1]|}{5770}  \\\hline
Unique words: UW & 611829 & 148073 & 167612 & 42104 & 175369 & 44169  \\\hline
Unique lemmas: UL & 282867 & 63351 & 70210 & 19698 & 75214 & 20637  \\\hline
Common UL & \multicolumn{2}{S[table-format=5.1]|}{60992} & \multicolumn{2}{S[table-format=5.1]|}{19138} & \multicolumn{2}{S[table-format=5.1]|}{20098}  \\\hline
Min words & 28 & 15 & 191 & 18 & 357 & 18 \\\hline
Max words & 1500 & 85 & 1500 & 85 & 1498 & 85 \\\hline
Avg words & 766.5 & 48.8 & 772.4 & 54.5 & 750.3 & 53.2 \\\hline
Avg sentences & 37.2 & 2.7 & 38.5 & 3.0 & 37.0 & 2.9  \\\hline
Avg UW & 419.1 & 41.3 & 424.2 & 46.0 & 415.7 & 45.1 \\\hline
Avg UL & 350.0 & 40.2 & 352.5 & 44.6 & 345.4 & 43.9 \\\hline
\end{tabular}
\end{table}

\begin{figure}
\includegraphics[width=\textwidth]{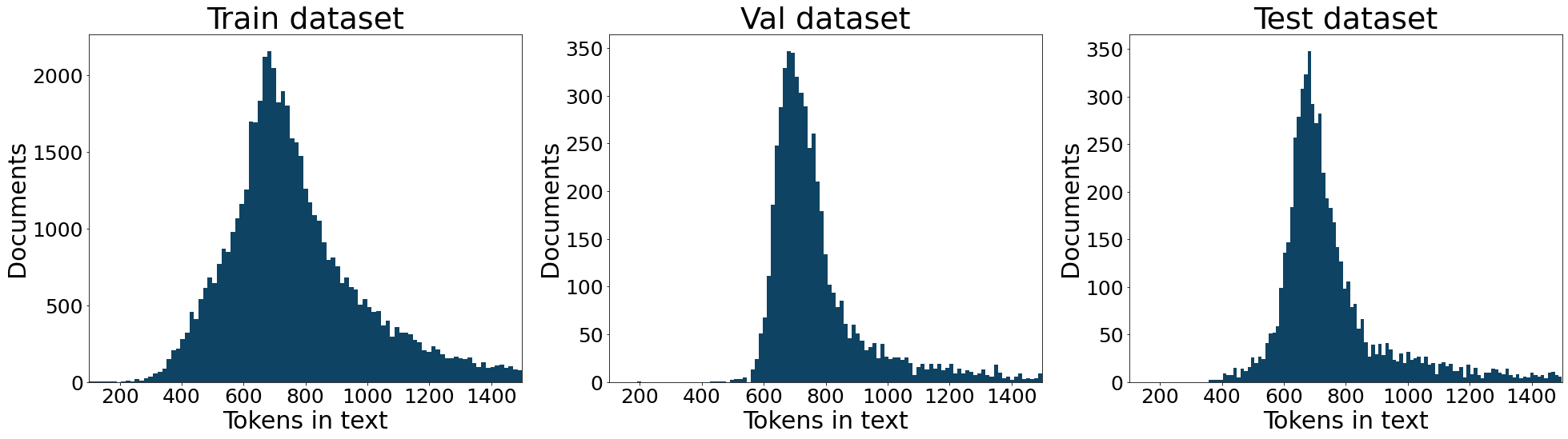}
\caption{Documents distribution by count of tokens in a text}
\label{fig1}
\end{figure}

\begin{figure}
\includegraphics[width=\textwidth]{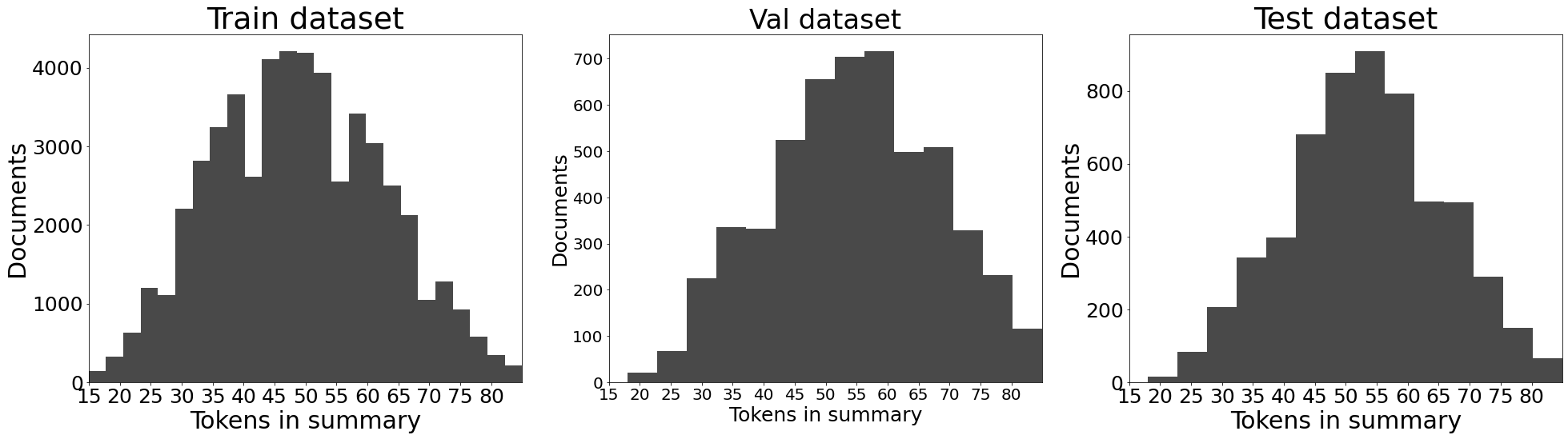}
\caption{Documents distribution by count of tokens in a summary} \label{fig2}
\end{figure}

We depict the distribution of tokens counts in texts in Figure~\ref{fig1}, and the distribution of tokens counts in summaries is in Figure~\ref{fig2}. The training dataset has a smoother distribution of text lengths in comparison with validation and test datasets. It also has an almost symmetrical distribution of summaries' lengths, while validation and test distributions are skewed.

To evaluate the dataset's bias towards extractive or abstractive methods, we measured the percentage of novel n-grams in summaries. Results are presented in Table 2 and show that more than 65\% of summaries' bi-grams do not exist in original texts. This number decreases to 58\% if we consider different word forms and calculate it on lemmatized bi-grams. Although we can not directly compare these numbers with CNN/DailyMail or any other English dataset as this statistic is heavily language-dependent, we should state that it is 53\% for CNN/DailyMail and 83\% for XSum. From this, we can conclude that the bias towards extractive methods can exist.

Another way to evaluate the abstractiveness is by calculating metrics of oracle summaries (the term is defined in \ref{sem}). To evaluate all benchmark models, we used ROUGE~\cite{rouge} metrics. For CNN/DailyMail oracle summaries score 31.2 ROUGE-2-F~\cite{presumm}, and for our dataset, it is 22.7 ROUGE-2-F.

\begin{table}[htbp]
\caption{Average \% of novel n-grams}\label{tab2}
\centering
\begin{tabular}{|c|C{1.4cm}|c|c|}\hline
& Train & \phantom{ha} Val \phantom{ck} &\phantom{ha} Test \phantom{ck}\\\hline
Uni-grams & 34.2 & 30.5 & 30.6 \\\hline
Lemmatized uni-grams & 21.4 & 17.8 & 17.6\\\hline
Bi-grams & 68.6 & 65.0 & 65.5 \\\hline
Lemmatized bi-grams & 61.4 & 58.0 & 58.5 \\\hline
Tri-grams & 84.5 & 81.5 & 81.9 \\\hline
\end{tabular}
\end{table}

\subsection{BPE}
We extensively utilized byte-pair encoding (BPE) tokenization in most of the described models. For Russian, the models that use BPE tokenization performs better than those that use word tokenization as it enables the use of rich morphology and decreases the number of unknown tokens. The encoding was trained on the training dataset using the sentencepiece~\cite{sentencepiece} library.

\subsection{Lowercasing}
We lower-cased all texts and summaries in most of our experiments. It is a controversial decision. On the one hand, we reduced vocabulary size and focused on the essential properties of models, but on the other hand, we lost important information for a model to receive. Moreover, if we speak about our summarization system's possible end-users, it is better to generate summaries in the original case.

We provide a non-lower-cased version of the dataset as the main version for possible future research.

\section{Benchmark methods}
We used several groups of methods. TextRank~\cite{textrank} and LexRank~\cite{lexrank} are fully unsupervised extractive summarization methods. Summarunner~\cite{summarunner} is a supervised extractive method. PG~\cite{pg}, CopyNet~\cite{copynet}, mBART~\cite{mbart} are abstractive summarization methods.

\subsection{Unsupervised methods}
This group of methods does not have any access to reference summaries and utilizes only original texts. All of the considered methods in this group extract whole sentences from a text, not separated words.

\subsubsection{TextRank}
TextRank~\cite{textrank}  is a classic graph-based method for unsupervised text summarization. It splits a text into sentences, calculates a similarity matrix for every distinct pair of them, and applies the PageRank algorithm to obtain final scores for every sentence. After that, it takes the best sentences by the score as a predicted summary. We used TextRank implementation from the summa~\cite{summa}\footnote{https://github.com/summanlp/textrank} library. It defines sentence similarity as a function of a count of common words between sentences and lengths of both sentences.

\subsubsection{LexRank}
Continuous LexRank~\cite{lexrank} can be seen as a modification of the TextRank that utilizes TF-IDF of words to compute sentence similarity as IDF modified cosine similarity. A continuous version uses an original similarity matrix, and a base version performs binary discretization of this matrix by the threshold. We used LexRank implementation from lexrank Python package\footnote{https://github.com/crabcamp/lexrank}.

\subsubsection{LSA}
Latent semantic analysis can be used for text summarization~\cite{lsa}. It constructs a matrix of terms by sentences with term frequencies, applies singular value decomposition to it, and searches right singular vectors' maximum values. The search represents finding the best sentence describing the k'th topic. We evaluated this method with sumy library\footnote{https://github.com/miso-belica/sumy}.

\subsection{Supervised extractive methods} \label{sem}

Methods in this group have access to reference summaries, and the task for them is seen as sentences' binary classification. For every sentence in an original text, the algorithm must decide whether to include it in the predicted summary.

To perform the reduction to this task, we first need to find subsets of original sentences that are most similar to reference summaries. To find these so-called ``oracle'' summaries, we used a greedy algorithm similar to SummaRunNNer paper~\cite{summarunner} and BertSumExt paper~\cite{presumm}. The algorithm generates a summary consisting of multiple sentences which maximize the ROUGE-2 score against a reference summary.

\subsubsection{SummaRuNNer}
SummaRuNNer~\cite{summarunner} is one of the simplest and yet effective neural approaches to extractive summarization. It uses 2-layer hierarchical RNN and positional embeddings to choose a binary label for every sentence. We used our implementation on top of the AllenNLP~\cite{allennlp}\footnote{https://github.com/allenai/allennlp} framework along with Pointer-Generator~\cite{pg} implementation.

\subsection{Abstractive methods}
All of the tested models in this group are based on a sequence-to-sequence framework. Pointer-generator and CopyNet were trained only on our training dataset, and mBART was pretrained on texts of 25 languages extracted from the Common Crawl. We performed no additional pretraining, though it is possible to utilize Russian headline generation datasets here.

\subsubsection{Pointer-generator}
Pointer-generator~\cite{pg} is a modification of a sequence-to-sequence RNN model with attention~\cite{rnn_attention}. The generation phase samples words not only from the vocabulary but from the source text based on attention distribution. Furthermore, the second modification, the coverage mechanism, prevents the model from attending to the same places many times to handle repetition in summaries.

\subsubsection{CopyNet}
CopyNet~\cite{copynet} is another variation of sequence-to-sequence RNN model with attention with slightly different copying mechanism. We used the stock implementation from AllenNLP~\cite{allennlp}.

\subsubsection{mBART for summarization}
BART~\cite{bart} and mBART~\cite{mbart} are sequence-to-sequence Transformer models with autoregressive decoder trained on the denoising autoencoding task. Unlike the preceding pretrained models like BERT, they focus on text generation even in the pretraining phase.

mBART was pretrained on the monolingual corpora for 25 languages, including Russian. In the original paper, it was successfully used for machine translation. BART was used for text summarization, so it is natural to try a pretrained mBART model for Russian summarization.

We used training and prediction scripts from fairseq~\cite{fairseq}\footnote{https://github.com/pytorch/fairseq}. However, it is possible to convert the model for using it within HuggingFace's Transformers\footnote{https://github.com/huggingface/transformers}. We had to truncate input for every text to 600 tokens to fit the model in GPU memory. We also used <unk> token instead of language codes to condition mBART.

\section{Results}
\subsection{Automatic evaluation}
We measured the quality of summarization with three sets of automatic metrics: ROUGE~\cite{rouge}, BLEU~\cite{bleu}, METEOR~\cite{meteor}. All of them are used in various text generation tasks and are based on the overlaps of N-grams. ROUGE and METEOR are prevalent in text summarization research, and BLEU is a primary automatic metric in machine translation. BLUE is a precision-based metric and does not take recall into account, while ROUGE uses both recall and precision-based metrics in a balanced way, and METEOR weight for the recall part is higher than weight for the precision part.

All three sets of metrics are not perfect as we only have only one version of a reference summary for each text, while it is possible to generate many correct summaries for a given text. Some of these summaries can even have zero n-gram overlap with reference ones.

We lower-cased and tokenized reference and predicted summaries with Razdel tokenizer to unify the methodology across all models. We suggest to all further researchers to use the same evaluation script.

\begin{table}[htbp]
\caption{Automatic scores for all models on the test set}\label{tab3}
\centering
\begin{tabular}{|l|C{1cm}|C{1cm}|C{1cm}|c|c|}\hline
& \multicolumn{3}{c|}{ROUGE} & \multirow{2}{*}{BLEU\footnotemark} & \multirow{2}{*}{Meteor}\\\cline{2-4}
 & 1 & 2 & L & & \\\hline
Lead-1 & 27.6 & 12.9 & 20.2 & 5.3 & 18.6 \\
Lead-2 & 30.6 & 13.7 & 25.6 & 10.9 & 23.7 \\
Lead-3 & 31.0 & 13.4 & 26.3 & 10.8 & 26.0 \\
Greedy Oracle & 44.3 & 22.7 & 39.4 & 17.7 & 35.5 \\\hline
TextRank & 21.4 & 6.3 & 16.4 & 3.9 & 17.5 \\
LexRank & 23.7 & 7.8 & 19.9 & 6.2 & 18.1 \\
LSA & 19.3 & 5.0 & 15.0 & 3.6 & 15.2 \\\hline
SummaRuNNer & 31.6 & 13.7 & 27.1 & 11.5 & \textbf{26.0} \\\hline
CopyNet & 28.7 & 12.3 & 23.6 & 8.5 & 21.0 \\
PG small & 29.4 & 12.7 & 24.6 & 9.0 & 21.2 \\
PG words & 29.4 & 12.6 & 24.4 & 8.9 & 20.9 \\
PG big & 29.6 & 12.8 & 24.6 & 9.3 & 21.5 \\
PG small +coverage & 30.2 & 12.9 & 26.0 & 10.1 & 22.7 \\
Finetuned mBART & \textbf{32.1} & \textbf{14.2} & \textbf{27.9} & \textbf{12.4} & 25.7 \\\hline
\end{tabular}
\end{table}

\footnotetext[10]{BLEU scores in earlier versions of the paper were incorrect. They were describing character-level n-grams instead of word-level n-grams. We apologize for the inconvenience.}

We provide all the results in Table~\ref{tab3}. Lead-1, lead-2, and lead-3 are the most basic baselines, where we choose the first, the first two, or the first three sentences of every text as our summary. Lead-3 is a strong baseline, as it was in CNN/DailyMail dataset~\cite{pg}. The oracle summarization is an upper bound for extractive methods.

Unsupervised methods give summaries that are very dissimilar to the original ones. LexRank is the best of unsupervised methods in our experiments.

\begin{figure}
\includegraphics[width=\textwidth]{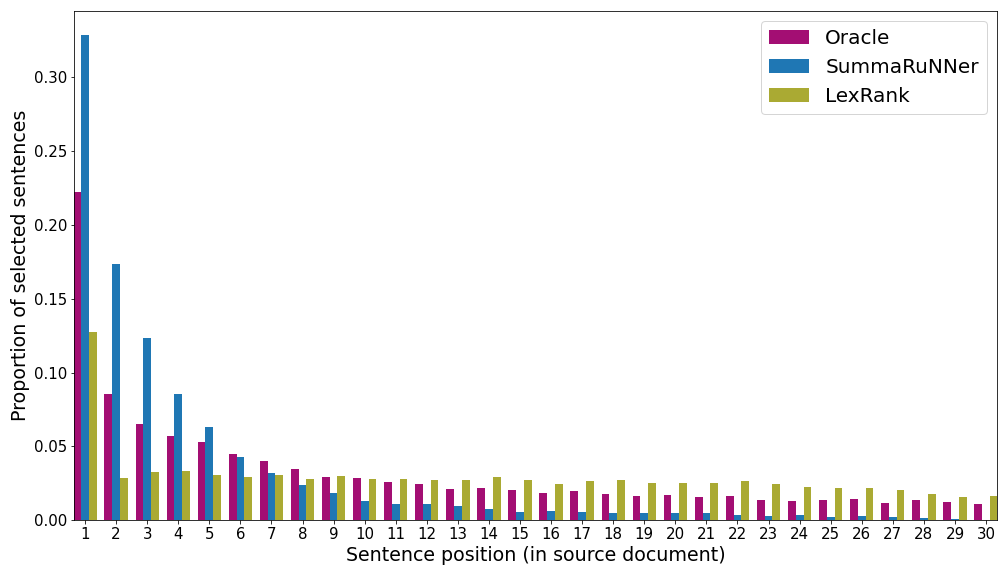}
\caption{Proportion of extracted sentences according to their position in the original document.}
\label{fig3}
\end{figure}

The SummRuNNer model has the best METEOR score and high BLEU and ROUGE scores. In Figure~\ref{fig3}, SummaRuNNer has a bias towards the sentences at the beginning of the text compared to the oracle summaries. In contrast, LexRank sentence positions are almost uniformly distributed except for the first sentence.

It seems that more complex extractive models should perform better on this dataset, but unfortunately, we did not have time to prove it.

To evaluate an abstractiveness of the model, we used extraction and plagiarism scores~\cite{ext_score}. The plagiarism score is a normalized length of the longest common sequence between a text and a summary. The extraction score is a more sophisticated metric. It computes normalized lengths of all long non-overlapping common sequences between a text and a summary and ensures that the sum of these normalized lengths is between 0 and 1.

\begin{figure}
\includegraphics[width=\textwidth]{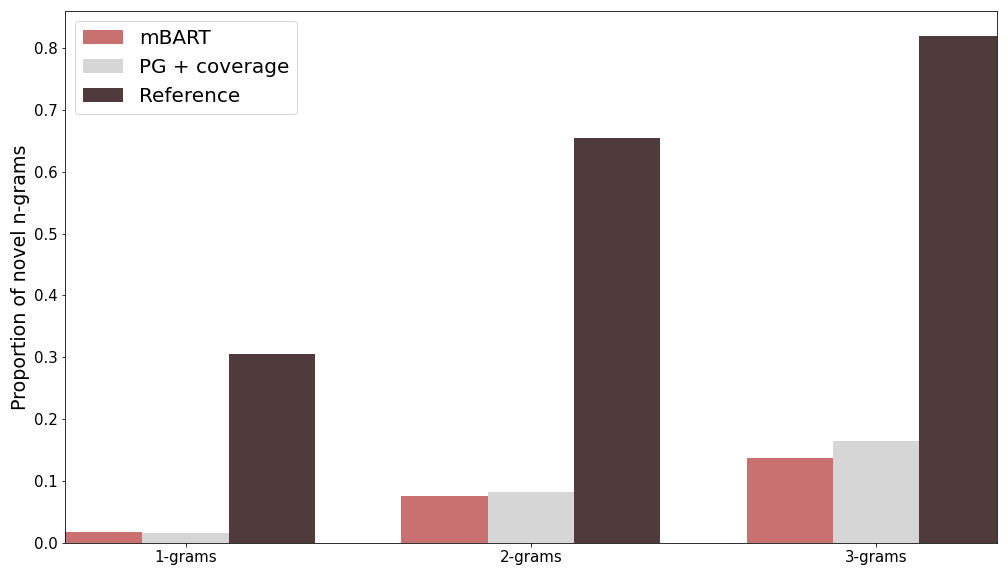}
\caption{Proportion of novel n-grams in model generated summaries on the test set} \label{fig4}
\end{figure}

As for abstractive models, mBART has the best result among all the models in terms of ROUGE and BLEU. However, Figure~\ref{fig4} shows that it has fewer novel n-grams than Pointer-Generator with coverage. Consequently, it has worser extraction and plagiarism scores~\cite{ext_score} (Table~\ref{tab4}).

\begin{table}[htbp]
\caption{Extraction scores on the test set}\label{tab4}
\centering
\begin{tabular}{|l|c|c|}\hline
& Extraction score & Plagiarism score\\\hline
Reference & 0.031 & 0.124\\
PG small +coverage & 0.325 & 0.501\\
Finetuned mBART & 0.332 & 0.502\\
SummaRuNNer & 0.513 & 0.662 \\\hline
\end{tabular}
\end{table}

\subsection{Human evaluation}
We also did side-by-side annotation of mBART and human summaries with Yandex.Toloka\footnote{https://toloka.yandex.ru/}, a Russian crowdsourcing platform. We sampled 1000 text and summary pairs from the test dataset and generated a new summary for every text. We showed a title, a text, and two possible summaries for every example. Nine people annotated every example. We asked them which summary is better and provided them three options: left summary wins, draw, right summary wins. The side of the human summary was random. Annotators were required to pass training, exam, and their work was continuously evaluated through the control pairs (``honeypots'').

\begin{wraptable}{r}{8cm}
\caption{Human side-by-side evaluation}\label{tab5}
\centering
\begin{tabular}{|c|c|c|}\hline
Votes for winner & Reference wins & mBART wins \\\hline
Majority & 265 & 735\\\hline
9/9 & 7 & 47 \\
8/9 & 18 & 106\\
7/9 & 30 & 185\\
6/9 & 54 & 200\\
5/9 & 123 & 180\\
4/9 & 32 & 17\\
3/9 & 1 & 0\\
\hline
\end{tabular}
\end{wraptable}

Table~\ref{tab5} shows the results of the annotation. There were no full draws, so we exclude them from the table. mBART wins in more than 73\% cases. We cannot just conclude that it performs on a superhuman level from these results. We did not ask our annotators to evaluate the abstractiveness of the summaries in any way. Reference summaries are usually too provocative and subjective, while mBART generates highly extractive summaries without any errors and with many essential details, and annotators tend to like it. The annotation task should be changed to evaluate the abstractiveness of the model. Even so, that is an excellent result for mBART.

Table~\ref{tab6} shows examples of mBART losses against reference summaries. In the first example, there is an unnamed entity in the first sentence, ``by them'' (``ими''). In the second example, the factual error and repetition exist. In the last example, the last sentence is not cohesive.

\begin{table}
\caption{mBART summaries that lost 9/9}\label{tab6}
\centering
\begin{tabular}{ | p{11.7cm} |}
\hline
разработанный ими метод идентификации способен выделить специфические для конкретного человека белки из пряди волос длиной всего сантиметра . это позволит с высокой степенью точности идентифицировать людей и без выделения днк .\\
\hline
\hline
\hline
президент россии владимир путин на встрече с ветеранами и представителями общественных патриотических объединений заявил , что каждый год единовременные выплаты ко дню победы составляют по 10 тыс . рублей ветеранам и по 5 тыс . рублей труженикам тыла . по 50 тыс . рублей также будет выплачено труженикам тыла . ранее в послании федеральному собранию президент также подчеркнул важность предстоящего юбилея вов .\\
\hline
\hline
\hline
самый одинокий актер голливуда , наконец , официально нашел пару . киану ривз , который многие годы предпочитал не распространяться о своей личной жизни и после давней трагедии решил не иметь детей , пришел на светское мероприятие с 46-летней художницей из лос-анджелеса александрой грант , чем вызвал ажиотаж у журналистов .', 'на арт-ивенте lacma art + film gala , прошедшем при поддержке gucci , актер киану ривз завел девушку — впервые за последние 20 лет . по словам артиста , в этом кругу редко вращается и ривз , несколько лет вызывающий сочувствие пользователей соцсетей фотографиями с празднований своего дня рождения .\\
\hline
\end{tabular}
\end{table}

\section{Conclusion}
We present the first corpus for text summarization in the Russian language. We demonstrate that most of the text summarization methods work well for Russian without any special modifications. Moreover, mBART performs exceptionally well even if it was not initially designed for text summarization in the Russian language.

We wanted to extend the dataset using data from other sources, but there were significant legal issues in most cases, as most of the sources explicitly forbid any publishing of their data even in non-commercial purposes.

In future work, we will pre-train BART ourselves on standard Russian text collections and open news datasets. Furthermore, we will try the headline generation as a pretraining task for this dataset. We believe it will increase the performance of the models.

\end{document}